\documentclass[10pt,letterpaper]{article}
\usepackage{natbib}
\usepackage{url}
\usepackage{pslatex}
\usepackage[utf8]{inputenc}
\usepackage{tabularx}
\usepackage{cogsci}
\usepackage{amsfonts}
\usepackage{booktabs}
\usepackage{blindtext}
\usepackage{pdflscape}
\usepackage{array}
\usepackage{etoolbox}
\usepackage{soul}
\usepackage{todonotes}
\usepackage[english]{babel}
\newcolumntype{L}[1]{>{\raggedright\let\newline\\\arraybackslash\hspace{2pt}}m{#1}}
\newcolumntype{C}[1]{>{\centering\let\newline\\\arraybackslash\hspace{2pt}}m{#1}}
\newcolumntype{R}[1]{>{\raggedleft\let\newline\\\arraybackslash\hspace{2pt}}m{#1}}

\usepackage{amsmath}

\usepackage{placeins}
\usepackage{graphicx}
\makeatother
\usepackage[font=small]{caption}  
\usepackage[justification=centering,font=scriptsize]{subfig} %for creating panels
\usepackage[countmax]{subfloat} %for creating panels

\usepackage{color}
\definecolor{Green}{RGB}{10,200,100}

\makeatletter
\def\hlinewd#1{%
\noalign{\ifnum0=`}\fi\hrule \@height #1 %
\futurelet\reserved@a\@xhline}
\makeatother

\title{Evaluating Compositionality in Sentence Embeddings}
 
\author{\textbf{Ishita Dasgupta$^1,$  Demi Guo$^2,$ Andreas Stuhlm\"uller$^3,$ Samuel J. Gershman$^4$ \& Noah D. Goodman$^3$}\medskip\\ 
$^1$Department of Physics and Center for Brain Science, Harvard University\\
$^2$Department of Computer Science, Harvard University\\
$^3$Departments of Psychology and Computer Science, Stanford University\\
$^4$Department of Psychology and Center for Brain Science, Harvard University
}

\begin{document}

\maketitle

\begin{abstract}
%we want to investigate compositionality in sentence embeddings - to what extent is it there in sota models, and if it’s not there, can these architectures represent it if trained with different data? comparisons are a particularly simple example of a case where you need to go beyond word meanings, so we start there. we find that infersent is bad at comparisons out of the box , but can learn to do well given appropriate training data. this is evidence that evaluation and teaching of compositional structure could work more generally using specifically constructed datasets.

% 150 word abstract
An important challenge for human-like AI is compositional semantics. Recent research has attempted to address this by using deep neural networks to learn vector space embeddings of sentences, which then serve as input to other tasks. We present a new dataset for one such task, “natural language inference” (NLI), that cannot be solved using only word-level knowledge and requires some compositionality. We find that the performance of state of the art sentence embeddings (InferSent; Conneau et al., 2017) on our new dataset is poor. We analyze the decision rules learned by InferSent and find that they are consistent with simple heuristics that are ecologically valid in its training dataset. Further, we find that augmenting training with our dataset improves test performance on our dataset without loss of performance on the original training dataset. This highlights the importance of structured datasets in better understanding and improving AI systems.

%An important frontier in the quest for human-like AI is compositional semantics: how do we design systems that understand an infinite number of expressions built from a finite vocabulary? Recent research has attempted to solve this problem by using deep neural networks to learn vector space embeddings of sentences, which then serve as input to supervised learning problems like paraphrase detection and sentiment analysis. Here we focus on ``natural language inference'' (NLI) as a critical test of a system's capacity for semantic compositionality. In the NLI task, sentence pairs are assigned one of three categories: entailment, contradiction, or neutral. We present a new set of NLI sentence pairs that cannot be solved using only word-level knowledge and instead require some degree of compositionality. We use state of the art sentence embeddings trained on NLI (InferSent, \cite{Conneau:2017uf}), and find that performance on our new dataset is poor, indicating that the representations learned by this model fail to capture the needed compositionality. We analyze some of the decision rules learned by InferSent and find that they are largely driven by simple heuristics at the word level that are ecologically valid in the SNLI dataset on which InferSent is trained. Further, we find that augmenting the training dataset with our new dataset improves performance on a held-out test set without loss of performance on the SNLI test set. This highlights the importance of structured datasets in better understanding, as well as improving the performance of, AI systems.\\
\textbf{Keywords:} 
Sentence embeddings; compositionality; test datasets
\end{abstract}

\section{Introduction}

A hallmark of human intelligence is compositionality: the ability, in the words of von Humboldt, to ``make infinite use of finite means.'' The failure of neural network models to achieve compositionality has been a recurring (and controversial) theme in cognitive science \citep{fodor88,gershman15,lake17}. However, recent successes of powerful deep learning systems trained on large corpora have renewed hopes that neural networks can close the gap with humans. In this paper, we explore minimal cases in a ``natural language inference'' (NLI, \cite{maccartney2009natural,dagan2006pascal}) task that cannot be solved without taking compositional information into account and thus develop a stringent test for compositionality. We then ask to what extent the state-of-the-art system for performing this task exhibits a truly compositional understanding of natural language. 

Our approach is motivated partly by the need for better benchmarks to assess AI systems 
%ettinger2016probing,
\citep{white2017inference,marelli2014sick,pavlick2016most,gershman15}. Currently, most systems are trained and evaluated on large corpora which can be partially gamed by simple heuristics. For example, \citet{socher11} presented a recursive autoencoder that achieved state-of-the-art performance on paraphrase detection, yet it only performed 10\% better than a baseline method that simply reported the most frequent class. The fact that these highly sophisticated algorithms may only be doing slightly better than naive baselines is brought into focus by more diagnostic benchmarks. %\citet{gershman15} showed that there is systematic structure in human phrase similarity judgments that is not captured by models such as the recursive autoencoder. 
We see a role for cognitive science in designing benchmarks that better probe the competences of AI systems, much in the same way that cognitive scientists have been probing the competences of humans \citep{ritter17,lake18}. 

Our results show that while the system we test exhibits poor performance on our compositional test set, much of its failure can be traced to biases in the training dataset. Furthermore, we see that the system is capable of exhibiting some compositionality given the right training data, pointing to potential uses for such structured datasets not just as diagnostic tools, but also for improving training of models.

%{TODO: i (ndg) think teh intro should provide a bit more detail about what we do in this paper: explore minimal cases of entailment tasks that can't be solved without taking compositional relations into account. Find that InferSent resoundingly fails. Track this failure down to biases in the SNLI training data, which may be inherrent in the collection method. Show that the InferSent architecture is nonetheless capabe of capturing these cases of composition when trained to do so....}

\section{Background}

\subsection{Sentence Embeddings}
Vector-based models of word semantics have been successful in capturing many aspects of word meanings. However, understanding language requires not only understanding words, but understanding their relations within a sentence. 
Due to the combinatorial productivity of language, the number of possible sentences far exceeds the size of the vocabulary; therefore generating similar vector embeddings for sentences has proven challenging. Recent literature reports several supervised as well as unsupervised approaches to learning sentence representations using Recurrent neural networks (RNNs) that account for word ordering \citep{kiros2015skip, Hill:2016uu, Conneau:2017uf}. 
% Consider commenting on neural Turing machine type thing - i.e. nn with memory (in theory much more flexible than RNN)?
These are intended to capture semantic content, and do perform reasonably well on transfer tasks---i.e. other sentence-level tasks which the embeddings were not specifically trained on. Particularly, the performance of these sentence models exceeds the performance of bag-of-words models that patently lack any relational information about the words (i.e., compositionality). However, it is unclear exactly what compositional information is gained in RNN sentence models beyond lexical meaning. 

\subsection{Natural Language Inference classifiers}
\begin{figure}[ht!]
\centering
\includegraphics[width=0.4\textwidth]{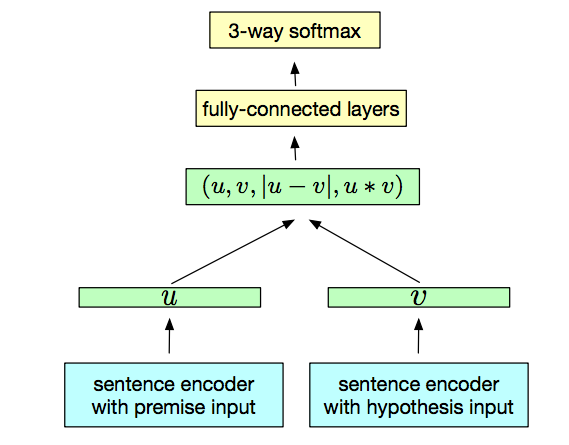}
\caption{InferSent architecture \citep{Conneau:2017uf}.}
\label{fig:arch}
\end{figure}

The sentence embeddings we explore in this paper are from InferSent \citep{Conneau:2017uf}. We choose to use these sentence embeddings as they represent the current state-of-the-art for transfer in semantic tasks, and we expect that strong performance in transfer tasks indicates a good representation for the semantics of a sentence. These embeddings were trained end-to-end using the architecture in Figure \ref{fig:arch} on the SNLI (Stanford Natural Language Inference) training set \citep{snli:emnlp2015}. The training task is to classify pairs of sentences into `entailment', `contradiction', or `neutral'. The embeddings were shown to perform well on other tasks (such as sentiment analysis, semantic textual similarity and other natural language inference datasets) by re-using the embedding layers and training only the classifier for the specific task at hand. We train the model using the same protocol as in \citet{Conneau:2017uf} for use in this work. Our trained InferSent model gives us 84.73\% accuracy on validation and 84.84\% accuracy on the SNLI test set, which is comparable the performance of the classifier reported in \citet{Conneau:2017uf}. For comparison, we also train a bag of words (BOW) baseline model that  averages the GloVe embeddings \citep{pennington14} for all the words in the sentence to form a sentence embedding. We train a multi-layer perceptron on these embeddings to give the BOW-MLP classifier we use in the following. BOW-MLP achieves $53.99\%$ accuracy on the SNLI test set \citep[comparable to the BOW performance reported in][]{Conneau:2017uf}.

% currently a bit degenerate with the abstract - maybe not req?
%\section{Contributions}
%We address the question of what is actually being learned by these sentence embeddings, by designing a diagnostic test dataset (based on comparisons) that relies explicitly on compositional information, and is intractable for systems that capture only lexical information. We characterize the performance of InferSent on this dataset. Given the exponentially large number of possible sentences, most subsets of sentences unless meticulously designed, can be justified by heuristics simpler than the true generative process of language. Performance of InferSent on the diagnostic dataset reveals that InferSent does seem to use some such heuristics when doing Natural Language Inference (NLI). This highlights the value of such structured test datasets in investigating and understanding what exactly some of these complex models encode. We also investigate the ecological validity of these heuristics in the SNLI dataset it is trained on. Finally, we retrain a model on a dataset containing both our comparison-based dataset and the original SNLI train set to see if structured datasets like ours can be learned by the InferSent architecture. This helps determine if the shortfall in compositionality found in InferSent is primarily due to poverty in the training data, or in the model architecture itself and also helps determine future utility of such structured datasets in improving training for models.

\subsection{SNLI dataset}
The Stanford Natural Language Inference dataset \citep{snli:emnlp2015} is a large annotated corpus for NLI that is generated with a crowdsourcing framework. Workers are presented with a scene description from a corpus of image captions, and asked to supply sentences that have each of three possible relations (entailment, neutral, and contradiction) to the given sentence. The freedom to produce entirely novel sentences leads to a rich set of examples from the set of possible sentences; however, it also leads to some unexpected biases that we will discuss in later sections.

\subsection{Notion of Compositionality}
Compositionality can mean many things. The notion that we focus on for this work is the abstract understanding of how words combine, in a way that generalizes to words and phrases that  have not previously been encountered. For example, rules of the type in Table \ref{tab:rules} hold true for X, Y and Z that may never have been encountered in that combination before. In fact, it should generalize to X, Y and Z that have never been encountered before at all. Understanding this sort of abstract rule, for any combinatorially large possible values for X, Y and Z, is a step to a more general understanding of compositional representations of sentence structure. %A more general notion arises from being able to combine and compose such abstract rules. In this paper, we will first deal with how to evaluate the learning of abstract rules, and if there are simple ways to encourage sentence representations to encode them. The abilities of such systems to compose such simple rules into more general ones is left to future work,

\section{Compositional comparisons dataset}
Our goal is to design pairs of sentences such that the NLI relation within a pair (entailment, neutral or contradiction) can be changed without changing the words involved, simply by changing the word ordering within each sentence. 
We thus generate sets of sentence pairs which differ by permutation of words, such that the pairs represent different relations.

\begin{table}[htb]
  %%\centering
  \resizebox{0.99\columnwidth}{!}{%
  \begin{tabular}{|| p{17mm} |p{35mm}| p{37mm} | p{14mm}||} 
 \hline
 Type & Entailment hypothesis & Contradiction hypothesis & \# of pairs\\
 \hline
    \hline
Same & 
X is more Y than Z  &
 Z is more Y than X & 14670\\
 \hline
More-Less &
Z is less Y than X  &
X is less Y than Z  & 14670\\
 \hline
Not &
Z is not more Y than X  &
X is not more Y than Z  & 14670\\
\hline
\end{tabular}}
 \caption{Comparisons dataset summary.
 Set of rules for premise: X is more Y than Z}
  \label{tab:rules}
\end{table}

%For instance, ``The woman is more cheerful than the man'' CONTRADICTS ``The woman is less cheerful than the man'', but ``The woman is more cheerful than the man'' ENTAILS ``The man is less cheerful than the woman.''

By construction, BOW models will perform at chance on this task, since they cannot distinguish the pairs. This provides a hard baseline for the performance that is possible without abstract rule understanding. In the literature, any performance above a BOW model is often seen as proof of compositionality. However, this is an unwarranted conclusion---the BOW model baseline usually receives only averaged word vectors for the sentence and therefore theoretically also loses some of the lexical information.  We propose to instead gauge the compositionality of sentence-vector models by seeing how differently they classify these permuted sets. 

We generate our test dataset using comparisons, as these yield many simple examples of sentence pairs that require more than word-level data to understand (when comparing two entities, their order in the sentence matters), and generation of several such sentence pairs can be easily automated. We consider three sub-types, described below and summarized in Table \ref{tab:rules}.

\subsection{Same type}
A-B pairs differ only in the order of the words. \\
{\tt A: The woman is more cheerful than the man \\ B: The man is more cheerful than the woman \\ CONTRADICTION \\} {\tt A: The woman is more cheerful than the man \\ B: The woman is more cheerful than the man\\ ENTAILMENT}

\subsection{More-Less type}
A-B pairs differ by whether they contain the word `more' or the word `less'. \\
{\tt A: The woman is more cheerful than the man \\ B: The woman is less cheerful than the man\\ CONTRADICTION \\}{\tt A: The woman is more cheerful than the man \\ B: The man is less cheerful than the woman \\ ENTAILMENT} 

\subsection{Not type}
A-B pairs differ by whether they contain the word `not'. \\
{\tt A: The woman is more cheerful than the man \\ B: The woman is not more cheerful than the man\\ CONTRADICTION \\}{\tt A: The woman is more cheerful than the man \\ B: The man is not more cheerful than the woman \\ ENTAILMENT} 

% \begin{table}[htb]
%   %%\centering
%   \resizebox{0.9\columnwidth}{!}{%
%   \begin{tabular}{||p{30mm} c||} 
%  \hline
%  Type & Number of sentence pairs \\ [0.5ex] 
%  \hline\hline
%   Comparisons (same)  &  14670\\
%  \hline
%  Comparisons (more/less)  &  14670\\
%  \hline
%  Comparisons (not)  &  14670\\
%  \hline
% \end{tabular}}
%  \caption{Comparisons dataset summary.}
%   \label{tab:dataset}
% \end{table}

To facilitate comparison with the SNLI dataset, we ensure that the vocabulary distribution of the Comparisons dataset is similar to the original SNLI training dataset. Only a few words differ by more than $1 \%$ from their occurrence rate in SNLI, such as \textit{not, a, than, the, is, less, more}. This is inevitable given the general structure of the comparison sentence pairs we use. 

%\todo{Update this table: update this with accuracy on three set of data }

\section{Classification Analysis}

The overall performance of each of the classifiers on the Comparisons dataset are given in Table \ref{tab:pretrainperf}.

\begin{table}[htb]
  %%\centering
  \resizebox{0.9\columnwidth}{!}{%
  \begin{tabular}{||p{20mm}| p{20mm}| p{20mm}||} 
 \hline
 Type & BOW-MLP & InferSent \\
    \hline\hline
same & 50.0 &50.37\\
 \hline
 more/less & 30.24 &50.35\\
\hline 
not & 48.98 & 45.24\\
\hline
\end{tabular}}
 \caption{Performance on the Comparisons dataset.}
  \label{tab:pretrainperf}
\end{table}

\subsection{BOW-MLP}
\begin{figure}[ht!]
\centering
\includegraphics[width=0.30\textwidth]{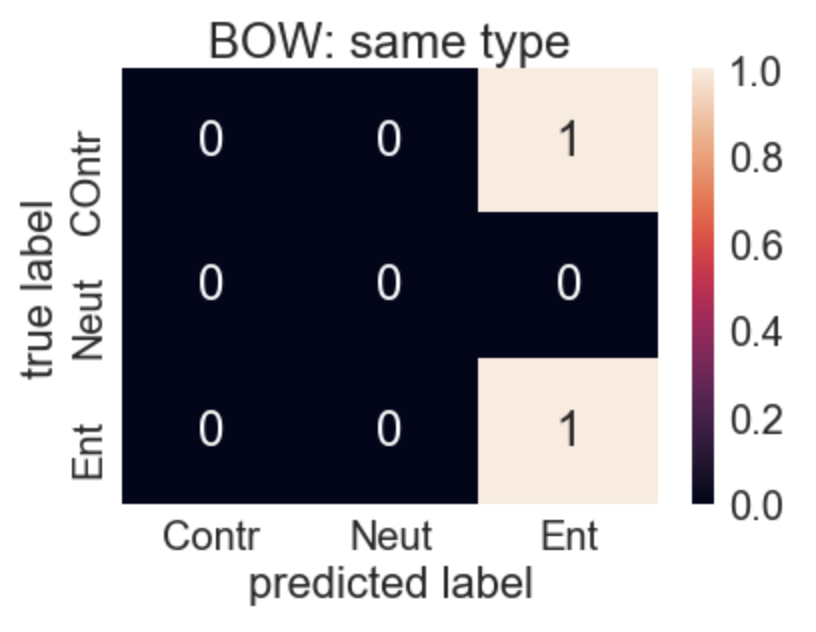}
\includegraphics[width=0.30\textwidth]{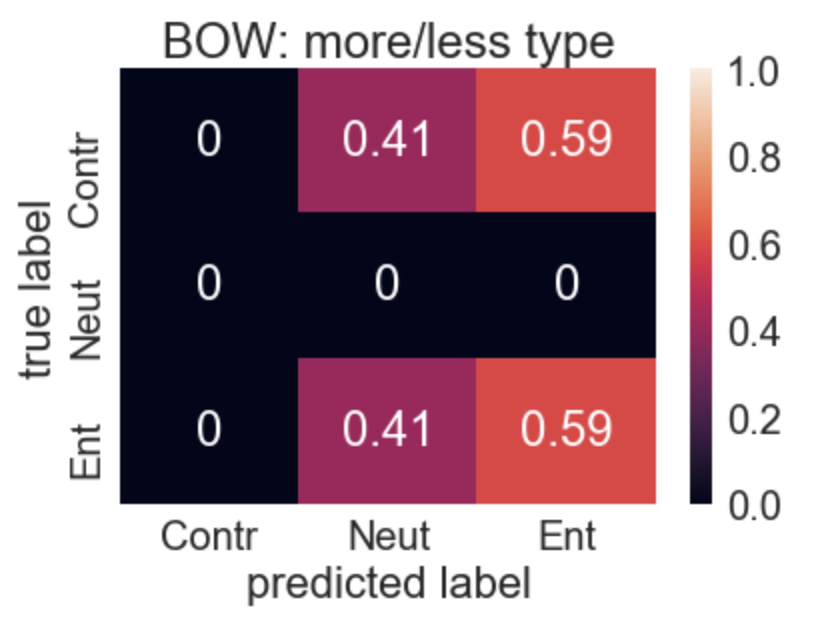}
\includegraphics[width=0.30\textwidth]{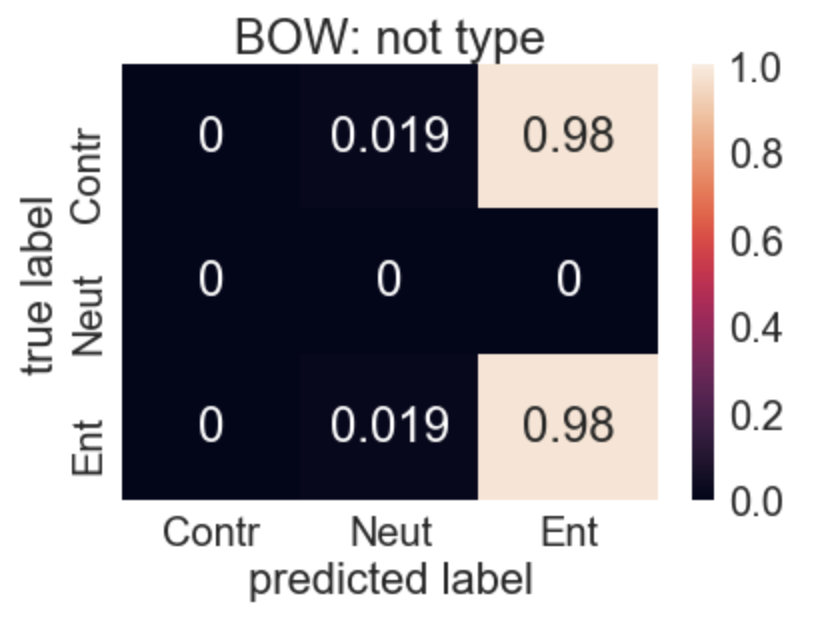}

\caption{BOW-MLP confusion matrices, with normalized rows.}

\label{fig:BOWhist}
\end{figure}

As expected, BOW-MLP makes classifications that are exactly symmetric across the two true categories in each task, since members of each category are just permuted versions of each other and BOW cannot distinguish them (Figure \ref{fig:BOWhist}). This also ensures that the performance is capped at 50\%. A sign of using more than word-level information would be asymmetry between the classifications of the two categories.

\begin{figure}[ht!]
\centering
\includegraphics[width=0.30\textwidth]{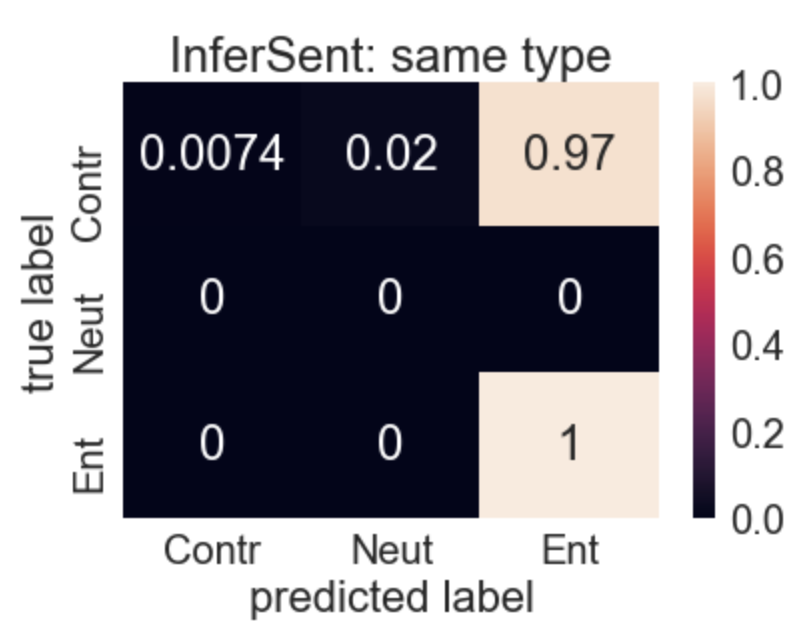}
\includegraphics[width=0.30\textwidth]{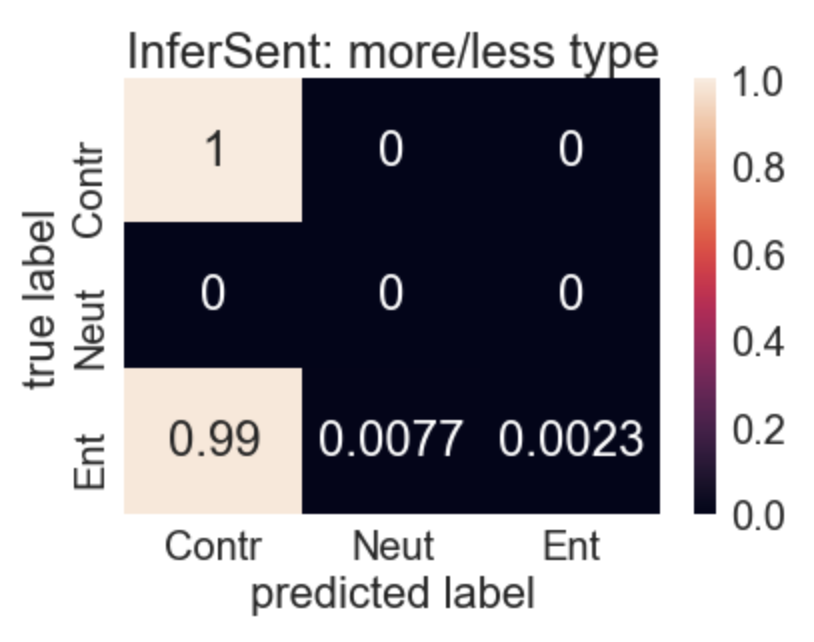}
\includegraphics[width=0.30\textwidth]{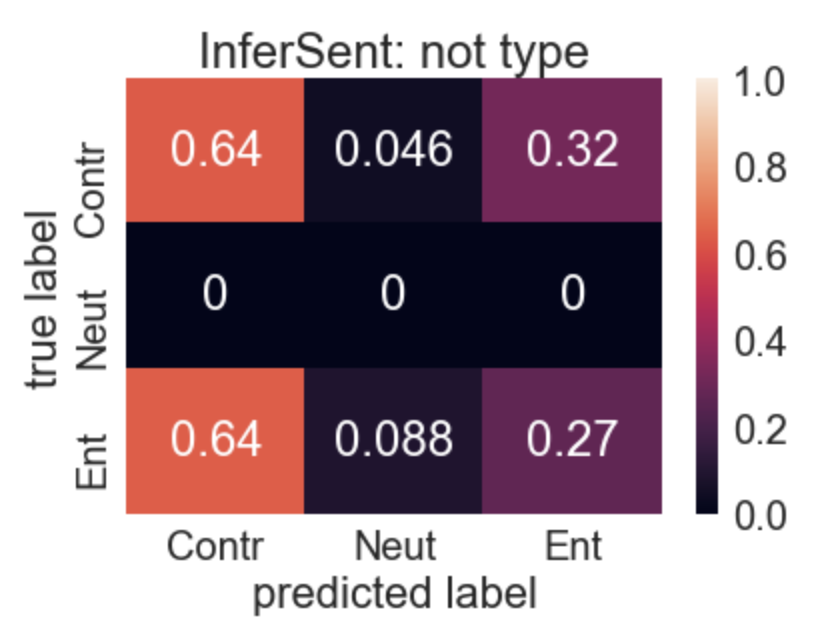}

\caption{InferSent confusion matrices, with normalized rows.}
\label{fig:InferSenthist}
\end{figure}

\subsection{InferSent}
The performance of InferSent is slightly more asymmetric (Figure \ref{fig:InferSenthist}), indicating that it is able to use some information beyond the word level. Yet overall InferSent is extremely poor at this task, indicating that it fails to fully exhibit the compositionality needed for these comparison sentences. We next analyze some of the patterns of classification errors observed.

\subsubsection{All same words}
When the words in both sentences are the same (the same-type comparisons) they are largely classified as entailments  (Figure \ref{fig:InferSenthist}), despite half being true contradictions. 
We observe that in the SNLI dataset, most contradictory sentence pairs have no overlap in words. For example, a contradictory sentence pair in SNLI is:\\
{\tt A: Several people are trying to climb a ladder in a tree. \\ B: People are watching a ball game. \\ CONTRADICTION\\}

%
% \begin{table}[htb]
%   %%\centering
%   \resizebox{\columnwidth}{!}{%
%   \begin{tabular}{|l|l|l|l|}
%   \hline
%      Top &Entailment & Neutral & Contradiction \\
%     \hline\hline
% All & 183416 (33.4 \%) & 182764 (33.3 \%) & 183187 (33.3 \%) \\
% \hline 
% 10000 & 3954 (39.5 \%) & 3567 (35.7 \%) & 2479 (24.8 \%)\\
% \hline
%  1000 & 508 (50.8 \%) & 407 (40.7 \%) & 85 (8.5 \%)\\
%  \hline
%   \end{tabular}}
%  \caption{High Overlap of words in SNLI.}
%   \label{tab:highOverlap}
% \end{table}

Thus, within SNLI, it is much more likely for a sentence pair to be entailment or neutral if they have significant overlap.
In order to quantitatively verify this observation, we rank all the sentence pairs by overlap rate: $\frac{\text{\# of overlap words}}{\text{total \# of words}}$(in non-increasing order). We then look at the top X sentences with highest overlap. As shown in Table \ref{tab:highOverlap}, $91.5\%$ of the pairs with top 1000 maximum overlap between the sentences have the true label of either entailment or neutral, and are very rarely true contradiction.

\begin{table}[htb]
  %%\centering
  \resizebox{\columnwidth}{!}{%
  \begin{tabular}{|l|l|l|l|}
  \hline
     Top &Entailment & Neutral & Contradiction \\
    \hline\hline
All & 33.4\% & 33.3\% & 33.3\% \\
\hline 
10000 & 39.5\% & 35.7\% & 24.8\%\\
\hline
 1000 & 50.8\% & 40.7\% & 8.5\%\\
 \hline
  \end{tabular}}
 \caption{Overlap rate of words in SNLI.}
  \label{tab:highOverlap}
\end{table}

Thus, InferSent may be learning the heuristic that high overlap in words predicts entailment, rather than a compositional semantic representation. This explains the failure of InferSent to generalize to our same-type task.

\subsubsection{Difference of one word}

When the words in two sentences differ by just one word, the decision is largely based on whether those words have opposing meanings irrespective of the order of the words. We see this from performance on more-less type comparisons (Figure \ref{fig:InferSenthist}). Here the words across the pairs differ only in the presence or absence of the word `more' or `less'.
%For example:\\
%{\tt A: The woman is more cheerful than the man \\ B: The woman is less cheerful than the man\\ CONTRADICTION \\}{\tt A: The woman is more cheerful than the man \\ B: The man is less cheerful than the woman \\ ENTAILMENT} \\
Since the relation between the words `more' and `less' is largely contradictory, we hypothesize that their use in a pair of sentences leads the classifier to presume the sentences are contradictory, irrespective of the order of the words. 
%We also find that in the top 1000 most overlapping sentence pairs, $60.0 \%$ of sentence pairs that Infersent classified as contradiction contain antonyms, while only $10.8 \%$ sentence pairs classified as entailment do. 

We evaluate this hypothesis by investigating the statistics of antonyms in the SNLI dataset. To check whether a sentence pair (A,B) contains antonyms, we go through each word in sentence A, and consider all synonyms of that word, and consider all antonyms of those synonyms. Finally, we check if sentence B contains any of those antonyms. %These synonyms and antonyms are found using the NLTK WordNet. \citep{bird2004nltk} .

\begin{table}[htb]
  %%\centering
  \resizebox{0.96\columnwidth}{!}{%
  \begin{tabular}{|l|l|l|l|}
  \hline
    & P(Antonym $|$ X) & P( X $|$ Antonym) \\
    \hline\hline
    X = Contradiction & 12.2\% & 61.2\% \\
      X = Entailment &  3.5\% &  18.0\% \\
\hline 
  \end{tabular}}
 \caption{ Antonym word pair in the SNLI dataset}
  \label{tab:antonyms}
\end{table}

We observe that this heuristic is fairly consistent with the SNLI data. Table \ref{tab:antonyms} shows that the presence of antonyms strongly predicts a true label of contradiction in the SNLI dataset (61.2\% compared to chance at 33.3\%). We also see that a true contradiction predicts the presence of an antonym pair (12.2\%) more strongly than entailment does (3.5\%).

\begin{table}[htb]
  %%\centering
  \resizebox{0.96\columnwidth}{!}{%
  \begin{tabular}{|l|l|l|l|}
  \hline
    & P(Antonym $|$ X) & P( X $|$ Antonym) \\
    \hline\hline
    X = Contradiction & 43.5\% & 28.9\% \\
      X = Entailment &  8.7\% &  34.3\% \\
\hline 
  \end{tabular}}
 \caption{ Antonym word pair in high overlap subset of SNLI.}
  \label{tab:antonyms-ho}
\end{table}

Since most of our dataset contains a large amount of overlap between sentences A and B, we check for statistics of the high overlap set as well (Table \ref{tab:antonyms-ho}). Here, the trend of a true contradiction predicting the presence of an antonym pair (43.7\%) more strongly than entailment does (8.7\%), is more pronounced. However, the presence of an antonym pair no longer predicts the true label of contradiction at a high rate. This is likely due to the very low base rate of contradictions in the high overlap subset of SNLI, as compared to entailments.

Overall, these results suggest, again, that the underlying statistics of the SNLI dataset lead models, including InferSent, to ignore the order of words when solving our more/less-type task.

%We also find that the presence of antonyms that are most used in the SNLI dataset are more likely to indicate contradiction for InferSent. For example, the antonyms  women/men, boy/girl and black/white appear more frequently in the SNLI sentence pairs than antonyms like old/young, sitting/standing, and sentences that differ by the first set of antonym pairs are more likely to be classified as contradiction by InferSent, than sentences that differ by the second set of antonym pairs.

\subsubsection{Negations}

Comparisons that differ in the presence  or absence of the negation `not' are preferentially classified as contradictions (Figure \ref{fig:InferSenthist}). To verify that this heuristic is largely consistent with the SNLI dataset, we look at sentence pairs that contain ``negating N-grams'': no, not, n't. (By considering ``n't'', we will consider words such as ``don't'' or ``doesn't''.)
\begin{table}[htb]
  %%\centering
  \resizebox{0.96\columnwidth}{!}{%
  \begin{tabular}{|l|l|l|l|}
  \hline
    & P(Negation $|$ X) & P( X $|$ Negation) \\
    \hline\hline
    X = Contradiction & 3.3 \% & 58.4 \% \\
      X = Entailment &  1.1 \% &  20.0 \% \\
\hline 
  \end{tabular}}
 \caption{Negation in SNLI dataset.}
  \label{tab:negation}
\end{table}

We observe that a ``negation difference yields contradiction'' heuristic is consistent with the SNLI data. Table \ref{tab:negation} shows that the presence of a negation strongly predicts a true label of contradiction in the SNLI dataset (58.4\% compared to chance at 33.3\%). We also see that a true contradiction predicts the presence of an antonym pair (3.3\%) slightly more strongly than entailment does (1.1\%).

\begin{table}[htb]
  %%\centering
  \resizebox{0.96\columnwidth}{!}{%
  \begin{tabular}{|l|l|l|l|}
  \hline
    & P(Negation $|$ X) & P( X $|$ Negation) \\
    \hline\hline
    X = Contradiction & 1.3\% & 60.0\% \\
      X = Entailment &  0.1\% &  7.5\% \\
\hline 
  \end{tabular}}
 \caption{Negation in the high overlap subset of SNLI.}
  \label{tab:negation-ho}
\end{table}

We repeat the analysis for the top 10,000 of the high overlap set as well (Table \ref{tab:negation-ho}). Here, the presence of negation predicts a contradiction even more strongly than in the full dataset (despite the lower base rates of contradiction in this subset of the data), indicating a very strong basis for this heuristic within the high overlap subset of the SNLI dataset.
%Note that we also did same experiment on more broadly defined negation ngrams which includes Neither, None and etc, and we get similar results. We have $43.6\%$ negation pairs are contradiction, while $27.3\%$ negation pairs are entailment. $7 \%$ contradiction pairs have negation while only $4 \%$ entailment pairs have negation.

\subsubsection{Summary of heuristics}

We find evidence for a few heuristics that explain the bulk of the patterns seen in the performance of InferSent on our Comparisons dataset, all of which have ecological validity in the SNLI dataset. First, we find that a large overlap in words between two sentences leads InferSent to believe that they entail one another. Second, we see that the difference of one word between the two sentences, when the difference is an antonym or a negation, leads InferSent to classify them as contradictions irrespective of word order. Both of these illustrate a disproportionate dependence on lexical, rather than compositional meaning in InferSent.

%We conjecture that these effects are due to the presence of short N-grams that contradict each other (`more X'/`less Y', `is'/`is not'), generalizing the single-word heuristics described above.  

%[TODO: this section still doesn't work for me. i wonder if we should cut it?]

The analysis so far has highlighted word-level heuristics that InferSent might be using. Yet the confusion matrix results (Figure \ref{fig:InferSenthist}) show a slight asymmetry, indicating at least minor multi-word effects. One hypothesis is that larger deviations in the order of overlapping words, alone, leads InferSent to dissfavor entailment.
This is trivially true for same-type comparisons where the exact same word order results in an entailment inference, and different word order sometimes leads to other classifications (top row of the same-type comparisons in Figure \ref{fig:InferSenthist}). But in this case these are the correct classifications, so the heuristic is indistinguishable from full compositional reasoning. Critically, in the case of comparatives of the `not' type, pairs that differ more in the word order are in fact entailments. 
% For example:\\
% {\tt A: The woman is more cheerful than the man \\ B: The man is not more cheerful than the woman} \\
% Entail one another but the two sentences differ more in word order than the corresponding pair of sentences that contradict each other:\\
% {\tt A: The woman is more cheerful than the man \\ B: The woman is not more cheerful than the man} \\
We see that for this type of example, InferSent classifies true contradictions as entailments \textit{more} than it does true entailments ($p = 0.2e-11$).

This suggests, though certainly doesn't prove, a heuristic that differing word order in the presence of `order-promoting' words like `more' and 'less' like in our Comparisons dataset, disfavors entailment judgments. 
There are other simple uses of word order that could be in play: for instance, antomynic pairs of bigrams could generalize the single-word heuristics described above.
However, a systematic analysis of the effect of word order, and of the ecological validity of such heuristics, is challenging due to the combinatorial explosion in the number of possibilities. We leave a thorough investigation of this to future work.

%Further, we also find that while all comparisons of the `more-less' type are classified as contradictions (\todo{insert figure}), the system is more confident about the ones that are truly entailment (\todo{update this number for new balanced dataset} $1867/2400$ pairs) being contradictions, i.e. the ones that have a larger perturbation to word order. 

%This suggests that InferSent might be using some heuristic encodings for word order. However, a systematic analysis of the effect of word order, and of the ecological validity of such heuristics, is challenging due to the combinatorial explosion in the number of possibilities. We leave a thorough investigation of this to future work.

% From qualitative analysis, we observe that if common antonyms are swapped in order, then Infersent will likely classify this sentence pair as a contradiction. For example:
% {\tt A: Boy sitting with a girl on the bench in the park \\ B: Girl sitting with a boy on the bench in the park}
%Moreover, we observe that when the order of common antonyms are swapped, then it's highly likely this is a contradiction in SNLI training dataset.
% @Demi: verify using quantitative data that above claims are correct
% TODO: check how confident InferSent is on temporal ordering classifications

%[TODO: somewhere, perhaps intro, we should describe how SNLI was generated, and why that might lead to unexpected biases. And why those biases in turn might encourage a sentence model to get by without much compositionality.]

\section{Augmented training}

The foregoing results suggest that biases in the SNLI training data may be enough to lead a sentence encoding model to use simple non-compositional representations. This leaves open the question of whether architectures such as InferSent are capable of representing the relational features needed to succeed at our Compositional task. In this section, we explore this question by retraining the InferSent model on a combined dataset which includes both the Comparisons dataset and original SNLI training data.  This serves to test whether simple training on examples of the rules in Table \ref{tab:rules}, will enable InferSent to generalize these rules to X, Y and Z that it has previously never seen in that combination. This is a step towards gauging the compositionality of this sentence representation.

The training subset of our Comparisons dataset consists of 40k sentence pairs ($7\%$ of the 550k pair SNLI training set). Validation and test sets consist of 2000 sentence pairs each. There is no overlap between any of these sets.

\subsection{Fine-tuning}
We first tried initializing with the model trained on the SNLI dataset, and then training it on our new Comparisons dataset (using the same protocols used in \cite{Conneau:2017uf} to train InferSent). Results are shown in Table \ref{tab:finetuning}.
We observed that model performance on the SNLI data task degrades over the course of training (test accuracy went from 84.84 \% to 56.37 \%), though it remained higher than the random baseline of 33.3 \%. The final model, however, performs very well on the Comparisons dataset (99.8 \% test accuracy).

So while a decline in the performance of the model on SNLI points to over-fitting to the  data, we see that the model doesn't simply memorize the specific training data used from the Comparisons dataset, and does actually learn the compositional rules (as evidenced by high test and validation performance on the Comparisons dataset). This indicates that the InferSent model architecture is in theory able to learn such relational patterns, given the right training data.
% * <dasgupta.ishita@gmail.com> 2018-05-13T18:24:26.568Z:
%
% ^.
% * <dasgupta.ishita@gmail.com> 2018-05-13T18:24:23.584Z:
%
% ^.
\begin{table}[htb]
  %%\centering
  \resizebox{1.0\columnwidth}{!}{%
  \begin{tabular}{||l|l|l|l||}
  \hline
    %Epoch & Train(Comp) & Val (SNLI) & Val (New) & Test (New)\\
    Epoch & Train(Comp) & Test(Comp) & Test(SNLI)\\
    \hline
    0 & 47.81\% & 45.36\% & 84.84\%\\
    %0 & N/A & 84.73 & 46.56 & 45.36\\
    \hline
    13 & 99.91\% & 99.8\% & 56.37\%\\

   % 13 & 100.0 & 55.42 & 99.8 & 99.55\\
\hline 
  \end{tabular}}
 \caption{Experiment: Finetuning}
  \label{tab:finetuning}
\end{table}
\vspace{-4pt}

\subsection{Retraining}
To explore whether it is possible to perform well on both the Comparisons and SNLI datasets, 
we next trained a model from scratch on a training dataset that includes both SNLI and our Comparisons dataset, again using the same training protocol used in the original paper on InferSent \citep{Conneau:2017uf}. Results are shown in Table \ref{tab:retraining}.
%[TODO: is the epoch number used due to early stopping? otherwise how was it chosen?]
% The  InferSent protocol does have an early stopping criteria, but I don't think we hit it. The stopping here was sort of arbitrarily chosen – i.e. when further training didn't result in tangible improvement in training error. Do you think we should report more detail for now? 
The test accuracy on SNLI (84.96 \%) is comparable to the model trained only on SNLI (84.84 \%). Moreover, test accuracy on the Comparisons dataset (99.55 \%) is much higher than the model trained only on SNLI (45.36 \%). Thus we show that it is possible to train a model such that it has high performance on specially designed edge-cases like the Comparisons dataset, without loss of performance on the more general SNLI dataset.

\begin{table}[htb]
  %%\centering
  \resizebox{1.0\columnwidth}{!}{%
  \begin{tabular}{||l|l|l|l||}
  \hline
    %Epoch & Train(Combined) & Val (SNLI) & Val (New) & Test (New)\\
    Epoch & Train(Combined) & Test(SNLI) & Test(Comp)\\
    \hline
    12 & 90.99\% & 84.96\% & 100.00\%\\
    %126 & 126 &- 126 -&- 126\\
    %12 & ? & ? & ?\\
    %12 & ? & ? & ?\\
    %12 & 94.12 & 84.34 & 100.0 & 100.0\\
\hline 
  \end{tabular}}
 \caption{Experiment: Retraining From Scratch}
  \label{tab:retraining}
\end{table}

This result also verifies our previous hypothesis that the model learns the simplest ways to accommodate the training data: the main reason the InferSent model performs badly on the Comparisons dataset is that its training data licenses ``shortcut'' biases, not because of shortcomings in the model itself. This points to the benefits of understanding potential biases in training data and including specially designed data to correct them. %In future work, we hope to better understand if such training results in a more general understanding of sentences.

%[TODO: would be great (and easy?) to evaluate SentEval on the SNLI+Comparisons trained model...]
% Demi is looking into it – results are comparable to original InferSent

\section{Discussion}

This work highlights the inadequacy of mainstream tasks in truly testing if Natural Language Processing (NLP) models represent compositional structure beyond the word level. InferSent achieves high performance on the test set of the SNLI dataset, as well as several other transfer tasks, but fails on our Comparisons dataset. This indicates that it misses crucial aspects of the compositionality in sentence meaning. How then does the InferSent model succeed on SNLI? Analysis of the behavior of the model on our well-controlled dataset allowed us to conjecture some word-driven heuristics, many of which we found have ecological validity in the SNLI training data.  

%First, we see that the difference of one word between the two sentences, when the difference is an antonym or a negation, leads InferSent to classify them as contradictions irrespective of word order, illustrating a disproportionate dependence on lexical, rather than compositional meaning in InferSent. We find that this heuristic has ecological validity in the training set. Second, we find evidence for a heuristic where, given large overlap in words, bigger changes to word order between the two sentences pushes the classification of their relation away from entailment. So we see that the order of the words is sometimes detected, and taken into account for classification, but not in a systematically compositional way. Finding ecological validity for permutations in SNLI is challenging and we leave this to future work.

This points to the utility of carefully designed datasets both for testing models' representational abilities, as well as for better understanding what they have actually learned. This is especially useful for models with large parameter spaces and many local minima, where heuristic solutions can explain much of the variance in the training data. 

Elucidating the blind spots in a system's encoding of compositionality can then be utilized to improve it. We found that the InferSent model can be trained to perform better on our Comparisons comparisons without reducing performance on SNLI, by just including a part of the comparison dataset in the training data. 
This indicates that, for this case, the shortcoming is not in the model architecture, but rather in the poverty and biases of the training data.
By debiasing training corpora and augmenting them with minimal contrasting examples, we can move closer to a truly compositional encoding of language.

\section{Future Directions}
Our Comparisons dataset has the crucial property that, by construction, it cannot be solved with only word-level information. 
Building a more general Comparisons dataset with this property that extends beyond comparison-type sentences is an important direction for future research. Another clear direction is to assess how other models, such as SkipThought \citep{kiros2015skip}, perform on these problems, and explore the heuristics they might be covertly employing. 
Using techniques for generating interpretable explanations from classification decisions 
\citep[e.g.][]{ribeiro2016should}
could help to better understand the strengths and weaknesses of these models on diagnostic datasets; and in turn perhaps prescribe new training regimes. 

Further work on augmented training will be needed to better isolate the benefits of including specially designed data in training: do the results transfer to other tasks that require similar aspects of compositionality or even to more distant aspects of understanding beyond the word level?

\subsection{Acknowledgements}

We are grateful to Anatole Gershman and Tim O'Donnell for helpful discussions. ID is supported by Microsoft Research. SJG is supported by the Office of Naval Research (N00014-17-1-2984). NDG is supported by DARPA agreement number FA8750-14-2-0009, and a Sloan Foundation Research Fellowship. Code and data are available at: \url{github.com/ishita-dg/ScrambleTests}.

% 150 word abstract:
%An important frontier in the quest for human-like AI is compositional semantics: how do we design systems that understand an infinite number of expressions built from a finite vocabulary? Recent research has attempted to solve this problem by using deep neural networks to learn vector space embeddings of sentences, which then serve as input to supervised learning problems like paraphrase detection and sentiment analysis. Here we focus on one such task, “natural language inference” (NLI), where the goal is to classify sentence pairs to one of three categories: entailment, contradiction, or neutral. We present a new set of NLI sentence pairs that cannot be solved using only word-level knowledge and instead require some degree of compositionality. We use sentence embeddings trained on InferSent (Conneau et al., 2017), a state-of-the-art NLI task, and find that performance on our new dataset is poor, indicating that the representations learned by this model fail to capture the needed compositionality. We analyze some of the decision rules learned by InferSent and find that it is largely driven by simple heuristics at the word level that are ecologically valid in the SNLI dataset on which InferSent is trained. Further, we find that augmenting the training dataset with our new dataset improves performance on a held-out test set without loss of performance on the SNLI test set. This highlights the importance of structured datasets in better understanding, as well as improving the performance of, AI systems.

\bibliographystyle{apa-good}
\def\thebibliography#1{\section*{References}
\fontsize{8.5}{8.3}\selectfont
 \list
 {[\arabic{enumi}]}{\leftmargin \parindent
	 \itemindent -\parindent
	 \itemsep 0ex plus 1pt
	 \parsep 0.2ex plus 1pt minus 1pt
	 \usecounter{enumi}}
	 \def\newbrick{\hskip .11em plus .33em minus .07em}
	 \sloppy\clubpenalty4000\widowpenalty4000
	 \sfcode`\.=1000\relax}
\bibliography{library}

\begin{thebibliography}{17}
\expandafter\ifx\csname natexlab\endcsname\relax\def\natexlab#1{#1}\fi
\expandafter\ifx\csname url\endcsname\relax
  \def\url#1{{\tt #1}}\fi
\expandafter\ifx\csname urlprefix\endcsname\relax\def\urlprefix{URL }\fi

\bibitem[{Bowman et~al.(2015)Bowman, Angeli, Potts, \&
  Manning}]{snli:emnlp2015}
Bowman, S.~R., Angeli, G., Potts, C., \& Manning, C.~D. (2015).
\newblock A large annotated corpus for learning natural language inference.
\newblock In {\em Proceedings of the 2015 Conference on Empirical Methods in
  Natural Language Processing (EMNLP)\/}. Association for Computational
  Linguistics.

\bibitem[{Conneau et~al.(2017)Conneau, Kiela, Schwenk, Barrault, \&
  Bordes}]{Conneau:2017uf}
Conneau, A., Kiela, D., Schwenk, H., Barrault, L., \& Bordes, A. (2017).
\newblock {Supervised Learning of Universal Sentence Representations from
  Natural Language Inference Data}.

\bibitem[{Dagan et~al.(2006)Dagan, Glickman, \& Magnini}]{dagan2006pascal}
Dagan, I., Glickman, O., \& Magnini, B. (2006).
\newblock The pascal recognising textual entailment challenge.
\newblock In {\em Machine learning challenges. evaluating predictive
  uncertainty, visual object classification, and recognising tectual
  entailment\/}, (pp. 177--190). Springer.

\bibitem[{Fodor \& Pylyshyn(1988)}]{fodor88}
Fodor, J.~A., \& Pylyshyn, Z.~W. (1988).
\newblock Connectionism and cognitive architecture: A critical analysis.
\newblock {\em Cognition\/}, {\em 28\/}, 3--71.

\bibitem[{Gershman \& Tenenbaum(2015)}]{gershman15}
Gershman, S., \& Tenenbaum, J.~B. (2015).
\newblock Phrase similarity in humans and machines.
\newblock In {\em Proceedings of the 37th Annual Conference of the Cognitive
  Science Society\/}.

\bibitem[{Hill et~al.(2016)Hill, Cho, \& Korhonen}]{Hill:2016uu}
Hill, F., Cho, K., \& Korhonen, A. (2016).
\newblock {Learning Distributed Representations of Sentences from Unlabelled
  Data}.

\bibitem[{Kiros et~al.(2015)Kiros, Zhu, Salakhutdinov, Zemel, Urtasun,
  Torralba, \& Fidler}]{kiros2015skip}
Kiros, R., Zhu, Y., Salakhutdinov, R.~R., Zemel, R., Urtasun, R., Torralba, A.,
  \& Fidler, S. (2015).
\newblock Skip-thought vectors.
\newblock In {\em Advances in neural information processing systems\/}, (pp.
  3294--3302).

\bibitem[{Lake \& Baroni(2017)}]{lake17}
Lake, B.~M., \& Baroni, M. (2017).
\newblock Still not systematic after all these years: On the compositional
  skills of sequence-to-sequence recurrent networks.
\newblock {\em arXiv preprint arXiv:1711.00350\/}.

\bibitem[{Lake et~al.(2018)Lake, Ullman, Tenenbaum, \& Gershman}]{lake18}
Lake, B.~M., Ullman, T.~D., Tenenbaum, J.~B., \& Gershman, S.~J. (2018).
\newblock Building machines that learn and think like people.
\newblock {\em Behavioral and Brain Sciences\/}, {\em 40\/}.

\bibitem[{MacCartney(2009)}]{maccartney2009natural}
MacCartney, B. (2009).
\newblock {\em Natural language inference\/}.
\newblock Stanford University.

\bibitem[{Marelli et~al.(2014)Marelli, Menini, Baroni, Bentivogli, Bernardi,
  Zamparelli et~al.}]{marelli2014sick}
Marelli, M., Menini, S., Baroni, M., Bentivogli, L., Bernardi, R., Zamparelli,
  R., et~al. (2014).
\newblock A sick cure for the evaluation of compositional distributional
  semantic models.
\newblock In {\em LREC\/}, (pp. 216--223).

\bibitem[{Pavlick \& Callison-Burch(2016)}]{pavlick2016most}
Pavlick, E., \& Callison-Burch, C. (2016).
\newblock Most" babies" are" little" and most" problems" are" huge":
  Compositional entailment in adjective-nouns.
\newblock In {\em Proceedings of the 54th Annual Meeting of the Association for
  Computational Linguistics (Volume 1: Long Papers)\/}, vol.~1, (pp.
  2164--2173).

\bibitem[{Pennington et~al.(2014)Pennington, Socher, \& Manning}]{pennington14}
Pennington, J., Socher, R., \& Manning, C. (2014).
\newblock Glove: Global vectors for word representation.
\newblock In {\em Proceedings of the 2014 conference on empirical methods in
  natural language processing (EMNLP)\/}, (pp. 1532--1543).

\bibitem[{Ribeiro et~al.(2016)Ribeiro, Singh, \& Guestrin}]{ribeiro2016should}
Ribeiro, M.~T., Singh, S., \& Guestrin, C. (2016).
\newblock Why should i trust you?: Explaining the predictions of any
  classifier.
\newblock In {\em Proceedings of the 22nd ACM SIGKDD International Conference
  on Knowledge Discovery and Data Mining\/}, (pp. 1135--1144). ACM.

\bibitem[{Ritter et~al.(2017)Ritter, Barrett, Santoro, \& Botvinick}]{ritter17}
Ritter, S., Barrett, D.~G., Santoro, A., \& Botvinick, M.~M. (2017).
\newblock Cognitive psychology for deep neural networks: A shape bias case
  study.
\newblock In {\em International Conference on Machine Learning\/}, (pp.
  2940--2949).

\bibitem[{Socher et~al.(2011)Socher, Huang, Pennin, Manning, \& Ng}]{socher11}
Socher, R., Huang, E.~H., Pennin, J., Manning, C.~D., \& Ng, A.~Y. (2011).
\newblock Dynamic pooling and unfolding recursive autoencoders for paraphrase
  detection.
\newblock In {\em Advances in Neural Information Processing Systems\/}, (pp.
  801--809).

\bibitem[{White et~al.(2017)White, Rastogi, Duh, \&
  Van~Durme}]{white2017inference}
White, A.~S., Rastogi, P., Duh, K., \& Van~Durme, B. (2017).
\newblock Inference is everything: Recasting semantic resources into a unified
  evaluation framework.
\newblock In {\em Proceedings of the Eighth International Joint Conference on
  Natural Language Processing (Volume 1: Long Papers)\/}, vol.~1, (pp.
  996--1005).

\end{thebibliography}
\end{document}